\documentclass[pmlr,twocolumn,10pt]{jmlr} 

\mlhtrack{findings}

\newif\iffinal
\finaltrue  

\iffinal
    \ifmlhneedspmlr
      \jmlrvolume{XXX}
      \jmlryear{2026}
    \fi
    \ifmlhfindings \jmlrproceedings{}{ML4H 2026 - Findings Track}\fi
    \ifmlhdemo     \jmlrproceedings{}{ML4H 2026 - Demo Track}\fi
    \jmlrworkshop{Machine Learning for Health (ML4H) 2026}
\else
    \jmlrproceedings{}{Submitted to ML4H 2026: \mlhtrackname}
    \jmlrworkshop{Machine Learning for Health (ML4H) 2026}
\fi

\usepackage{booktabs}
\usepackage{amsmath}
\usepackage{listings}

\usepackage[switch]{lineno}

\theorembodyfont{\upshape}
\theoremheaderfont{\scshape}
\theorempostheader{:}
\theoremsep{\newline}

\title[Effects of LLM-Induced Note Bloat and Diagnostic Overcall on Downstream LLMs]{AI in the Clinical Documentation Loop: Effects of LLM-Induced Note Bloat and Diagnostic Overcall on Downstream LLMs}

\author{
    \Name{Jinghui Liu} \Email{jinghui.liu@csiro.au}\\
    \Name{Anthony Nguyen} \Email{anthony.nguyen@csiro.au}\\
    \addr Australian e-Health Research Centre, CSIRO
    \AND
    \Name{Taylan Gurgenci} \Email{t.gurgenci@uq.edu.au} \\
    \addr Mater Research Institute – University of Queensland \\
    \addr School of Medicine and Dentistry, Griffith University
}

\begin{document}

\maketitle

\ifmlhdemo\else
\begin{abstract}

    Large language models (LLMs) are increasingly used to process medical records for healthcare tasks such as clinical documentation. Because these tools also generate documentation, the relationship is now bidirectional: models both read from and write to the record. Understanding how LLMs behave when reading records that already contain LLM-authored content is therefore critical for their safe deployment.
    In this study, we examine how LLM-authored content in the medical record affects downstream LLMs that read the same record. We consider two types of LLM-induced artifacts: \textbf{bloat}, in which LLM-generated summaries accumulate in the record and change the length and provenance of its content over time; and \textbf{distortion}, in which LLMs subtly alter clinical findings rather than fabricating facts outright. As a case of distortion, we study \emph{diagnostic overcall}---documenting an unconfirmed differential diagnosis in the record---and test whether it propagates to the reader LLM's output.
    Across seven reader models from four families, we find evidence of both effects. First, LLM-generated summary notes modestly but consistently degraded downstream hospital-course generation compared with copy-forward and templated text of matched length. Second, injected overcalls reappeared in downstream outputs at rates that varied by reader model, without detectable changes in standard quality metrics.
    These findings show the need to monitor AI-generated content in medical records and to account for it when those records are later processed by AI.

\end{abstract}
\begin{keywords}
Large language models, clinical documentation, note bloat, summarization, text generation evaluation.
\end{keywords}
\fi

\ifmlhneedsstatements
\paragraph*{Data and Code Availability}
The experiments were conducted on the MIMIC-III dataset~\citep{Johnson2016-kd}, which is available through PhysioNet under credentialed access. 
All models were run locally, in accordance with its data use agreement.
We have provided all the prompts used in the study to support reproducibility.

\paragraph*{Institutional Review Board (IRB)}
This study does not require IRB approval.
\fi

\section{Introduction}
\label{sec:intro}

\begin{figure*}
    \includegraphics[width=\linewidth]{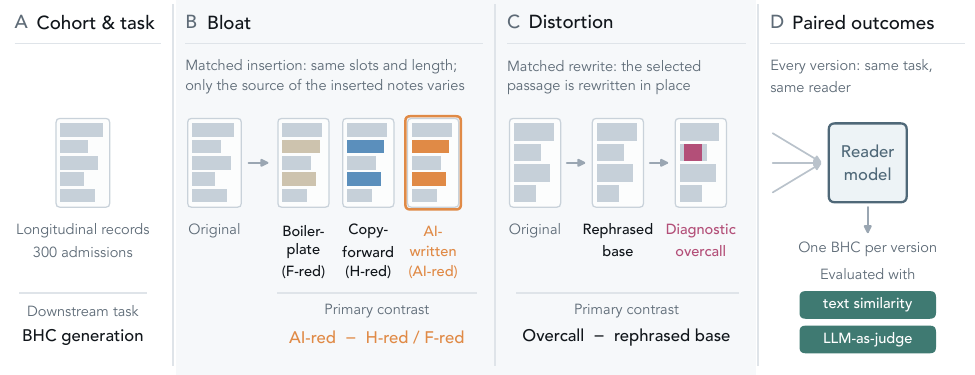} 
    \caption{Study design for controlled analyses of \textbf{bloat} and 
    \textbf{distortion} artifacts in AI-generated documentation and their 
    impact on downstream record readers. BHC: Brief Hospital Course, the hospital-course section of the discharge summary. 
    } 
    \label{fig:design}
\end{figure*}

AI-assisted documentation has attracted broad interest for its potential to help clinicians write clinical notes and relieve documentation burden. AI scribes, automated summarization, and discharge instruction generation are widely accessible and, in many cases, already implemented at scale~\citep{Tierney2025-vi,Ma2025-ni}. 
Meanwhile, the electronic health record (EHR) is increasingly being read and written by AI, which means that AI-generated content will be consumed by AI in later iterations to guide subsequent documentation and decision support.
This raises an important question: 
how are downstream LLM readers affected by AI-authored content already in the record?
Understanding how models behave under this dynamic is critical. Recent studies show that LLMs with strong medical QA capabilities may degrade substantially in real-world deployment, such as in interactive dialogue~\citep{Li2024-hs} or for user-driven assistance~\citep{Bean2026-ma}. 
How LLMs behave when processing real-world EHRs with noisy, LLM-authored context requires deeper investigation.

In this preliminary study, we consider two types of documentation artifacts potentially induced by AI-based documentation tools, and assess their effects on downstream models that read the records through controlled analyses. The first artifact is \textbf{``note bloat''}, where we perturb an original record by inserting AI-generated summaries of preceding notes and compare downstream performance on the perturbed and original records. 
The second involves \textit{distortion}, a nuanced form of error that we consider to be different from overt hallucination with fabricated facts. 
A distorted claim may contain truthful information stripped of context or incorrect information that remains hedged. For instance, a clinician may document as part of their impression that ``local recurrence (of cancer) unlikely'' not because the clinician thinks it is at all likely, but to guide other clinicians reading the note because the patient has asked about it many times. An AI model may interpret this as local recurrence being a differential being actively entertained. We consider this phenomenon as \textbf{``diagnostic overcall''}  in this study, referring to an unsupported differential documented by AI that clinicians would otherwise omit.
Figure~\ref{fig:design} illustrates the overall study design.

\section{Related Work}
\label{sec:related}

\paragraph{Reliability of LLMs in healthcare.}
LLMs achieve expert-level accuracy on exam-style medical questions~\citep{Singhal2025-ik} and reasoning tasks~\citep{Brodeur2026-we}, showing how rapidly the technology has progressed. However, LLMs may struggle to handle less curated questions or requests with noisy user inputs.
For example, when queried interactively rather than in a single turn, LLMs tend to degrade substantially in medical question-answering~\citep{Li2024-hs,Bean2026-ma}, a finding also corroborated in the general domain~\citep{Laban2026-wy}. Models could also struggle with misleading context, often abandoning previously correct answers~\citep{Zhou2026-vu}. The reliability of LLM-based documentation tools in real-world clinical settings therefore warrants careful evaluation.

\paragraph{Note bloat and LLMs.}
\citet{Liu2022-eu} studied how note bloat affects supervised NLP models, showing that it induces a data shift that degrades performance. 
\citet{Cahoon2026-ta} more recently proposed methods to identify and reduce repeated or redundant content in clinical notes for efficient LLM use. 
However, both works examined human-generated documentation artifacts rather than LLM-generated ones. Whether note authorship changes the effect remains an open question.

\paragraph{Long context and distraction.}
It is found that LLMs use information unevenly across long input context windows, with material in the middle of the context recovered least reliably~\citep{Liu2024-pa}. They can also be distracted by irrelevant but topically related content provided in the context~\citep{Shi2023-zg}. This shows the importance of separating the effects of context length and content provenance when analyzing note bloat.


\paragraph{Self-consuming loops.}
The recursive setting we describe has a close analogue in training dynamics, where models trained on their own outputs degrade across generations~\citep{Shumailov2024-hp,Alemohammad2024-yw}. 
However, a key distinction with these prior work is that they focus on training dynamics, whereas our setting concerns inference-time context, where the model consumes model-generated outputs. This is particularly relevant for clinical documentation where notes naturally accumulate over time, creating a recursive context that could amplify redundancies and errors. 


\section{Methods \& Experiments}

\begin{figure*}[t]
  \centering
  \includegraphics[width=\textwidth]{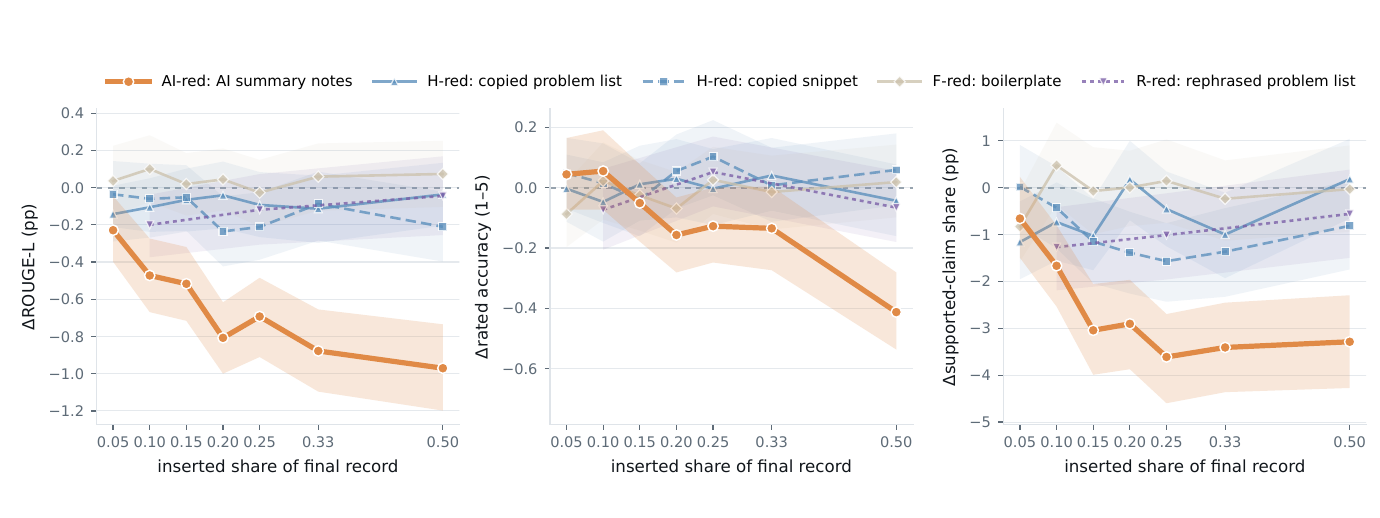}
  \caption{Effect of note bloat on downstream clinical summarization (main reader, Gemma-4-26B-A4B).
  X-axis: inserted share of the final record (dose).
  Y-axis: change from the same reader's
  output on the original record, in percentage points (pp) except rated accuracy (1--5 scale).
  Shades are 95\% bootstrap CIs. 
  }
  \label{fig:bloat-dose-main}
\end{figure*}

\subsection{Problem formulation}
\label{sec:formulation}
We study whether AI-generated content in a medical record changes what a downstream LLM (\emph{reader}) produces from that record. The downstream task is to write the Brief Hospital Course (BHC), the hospital-course section of the discharge summary. For admission $i$, the record $x_i$ contains all notes charted before discharge. A reader model $R$ receives these notes in chronological order, with timestamps and note types, and generates a BHC, $R(x_i)$; the reference $y_i^{\ast}$ is the clinician-written BHC from the discharge summary.
A documentation artifact, produced by a generator model $G$, transforms $x_i$ into a perturbed record that $R$ reads in the same way.
Every comparison is paired within an admission: a perturbed record is compared with a matched record that differs only in the artifact under study.

\subsection{Documentation artifacts}
\label{sec:artifacts}

\paragraph{Note bloat.}
Bloat inserts redundant notes into the record. We write $x_i^{a,d}$ for the record with notes of setting $a$ inserted at \textit{dose} $d$, which denotes the inserted fraction of the final record. At $d=0.5$, inserted text makes up half of $x_i^{a,d}$. We used seven doses from 0.05 to 0.50 and four settings: an AI summary of the preceding notes written by $G$ (\texttt{AI-red}); copied problem-oriented text or random snippets from preceding human-written notes (the two \texttt{H-red} variants); and templated boilerplate as filler (\texttt{F-red}).
Insertion positions were the same across settings, and \texttt{H-red} and \texttt{F-red} matched the length of \texttt{AI-red} at each position.
To separate authorship from content, $G$ also rewrote each copied problem-list note in its own words while keeping its content, giving a rephrased control for authorship (\texttt{R-red}, at doses 0.10, 0.25, and 0.50).
Appendix~\ref{app:bloat-construction} provides details on insertion and note generation.

\paragraph{Distortion with diagnostic overcall.}
Distortion edits an existing narrative note instead of inserting new ones. For admission $i$ and target type $t$, we chose a target diagnosis $c_{it}$ absent from the record. The generator $G$ first neutrally rephrased a selected passage, giving the control record $\tilde{x}_{it}$, then added $c_{it}$ to that rephrased passage as an unresolved, hedged differential, giving the distorted record $\tilde{x}_{it}^{+}$ (\texttt{ovc}). The two records differ only by the overcall, and both contain the same AI-rewritten passage, which holds AI authorship fixed.
Target types are defined by ICD code co-occurrence with the primary diagnosis: \texttt{ovc-related} targets frequently co-occur with it, and \texttt{ovc-unrelated} targets the opposite. Details of \texttt{ovc} filtering are provided in Appendix~\ref{app:ovc-construction}.

\subsection{Evaluation}
\label{sec:evaluation}
For bloat, summaries were scored against the reference by textual similarity (ROUGE-L and BERTScore, in percentage points or pp), and LLM-judged quality ratings and the share of supported claims scored by an LLM evaluator $E$ (Appendix~\ref{app:evaluation}). $s_i(x)=s\big(R(x),y_i^{\ast}\big)$ is the score of the summary of record $x$ under metric $s$. 
For distortion, $E$ compared the summaries of $\tilde{x}_{it}$ and $\tilde{x}_{it}^{+}$ and labeled $c_{it}$ in each as confirmed (\emph{hardened}), possible (\emph{carried}), or absent. We set $m_{it}(x)=1$ if $R(x)$ mentions $c_{it}$ as confirmed or possible.
The bloat effect and propagation rate are 
\begin{align}
  \Delta_d(a,b)&=\frac{1}{|\mathcal{I}|}\sum_{i\in\mathcal{I}}\big[s_i(x_i^{a,d})-s_i(x_i^{b,d})\big], \label{eq:bloat-effect}\\
  \pi_t&=\frac{1}{|\mathcal{I}_t|}\sum_{i\in\mathcal{I}_t} m_{it}(\tilde{x}_{it}^{+})\,\big[1-m_{it}(\tilde{x}_{it})\big], \label{eq:propagation}
\end{align}
where $a$ is the bloated record (e.g., \texttt{AI-red} or \texttt{H-red}), $b$ is the original record ($x_i^{b,d}=x_i$),
$\mathcal{I}$ is the set of admissions, and $\mathcal{I}_t$ includes those with target type $t$.
We report means with 95\% percentile-bootstrap intervals from 2,000 seeded resamples.

\subsection{Experimental setup}
\label{sec:setup}

\paragraph{Cohort.}
We used MIMIC-III~\citep{Johnson2016-kd}, which includes both progress notes and discharge summaries. We sampled 300 admissions, one per patient, after screening for record length, note composition, primary-diagnosis category, and the presence of hedged findings (Appendix~\ref{app:cohorts}).

\paragraph{Models.}
We used recent open-weight models, hosted locally with vLLM~\citep{Kwon2023-cu}. Qwen3.8-27B~\citep{Qiu2026-jl} was the generator $G$, and Nemotron-3.5-Lightning-30B-A3B~\citep{NVIDIA2026-nemotron35} served as an alternative $G$ for the main reader to test dependence on the generator. The main reader $R$ was Gemma-4-26B-A4B~\citep{Gemma-Team2026-fd}; six further readers spanned four model families: Gemma-4-12B, Gemma-4-31B, Nemotron-3.5-Lightning-30B-A3B, Llama-3.3-70B~\citep{Llama3-xd}, Llama-4-Scout-17B-16E~\citep{Meta2025-llama4}, and Qwen3.8-27B reading the notes it had generated ($R=G$). Qwen3.8-27B was also the evaluator $E$ but never scored its own summaries; GPT-OSS-120B was a second $E$ for distortion (Appendix~\ref{app:model-details}).

\begin{table}[t]
  \centering
  \small
  \setlength{\tabcolsep}{4pt}
  \begin{tabular}{@{}lcc@{}}
    \toprule
     & Related & Unrelated \\
    \midrule
    \multicolumn{3}{@{}l}{\textit{LLM evaluation: induced propagation (\%)}} \\
    Main reader & 24.6 {\scriptsize[16.8, 32.4]} & 16.1 {\scriptsize[10.9, 22.2]} \\
    \quad GPT-OSS evaluator & 23.8 {\scriptsize[16.8, 31.4]} & 15.6 {\scriptsize[10.2, 21.1]} \\
    \quad Nemotron generator & 10.2 {\scriptsize[5.1, 15.8]} & 8.3 {\scriptsize[4.4, 12.7]} \\
    Five other readers & 5.8--31.2 & 2.5--23.9 \\
    \midrule
    \multicolumn{3}{@{}l}{\textit{Clinician review (24 main-reader pairs)}} \\
    Evaluator agreement & \multicolumn{2}{c}{22/24 ($\kappa=0.83$)} \\
    Genuine overcalls & 6/12 & 8/12 \\
    \bottomrule
  \end{tabular}
  \caption{Diagnostic overcall propagation. Unless noted, Qwen3.8-27B wrote the overcalls and evaluated the summaries of the main reader (Gemma-4-26B-A4B). Brackets: 95\% intervals, Bonferroni-adjusted over the three main-reader rows. Results for other readers in Appendix Table~\ref{tab:readers}.}
  \label{tab:ovc-readers}
\end{table}

\section{Results \& Findings}

\subsection{AI-generated notes shifted downstream performance through their content}
\label{sec:results-ts-bloat}

Figure~\ref{fig:bloat-dose-main} shows the bloat effect (Eq.~\ref{eq:bloat-effect}) for the main reader. 
At the same inserted lengths and positions, AI-authored chart summaries produced
worse downstream BHCs than copied human text, as measured by textual similarity, LLM-judged quality, and verified claims. This effect increased with dose.
In contrast, equal-length \texttt{H-red} variants and \texttt{F-red} tended to remain similar to the original-record output. 
At dose 0.5, \texttt{AI-red} reduced judge-rated accuracy (1--5 scale) relative to snippet-based \texttt{H-red} and \texttt{F-red} by 0.47 (95\% CI: 0.34--0.59) and 0.43 points (0.31--0.55).
The effect is modest -- about half the run-to-run variation of a single summary -- but it is also consistent: the ROUGE-L reduction held for all seven readers ($-0.66$ to $-1.13$ pp at dose 0.25) and for notes from the alternative generator ($-0.57$ pp; Appendix~\ref{app:readers}).

The rephrased control showed no effect of AI authorship merely on its own: relative to the copied notes it rewrote, \texttt{R-red} changed ROUGE-L by $-0.07$, $-0.07$ and $0.00$ pp at doses 0.10, 0.25 and 0.50 (all intervals include zero), while \texttt{AI-red} still lowered ROUGE-L by 0.28, 0.59 and 0.92 pp against \texttt{R-red} (Appendix Table~\ref{tab:authorship}). The bloat effect thus comes from the AI-generated content rather than wording.

\subsection{Diagnostic overcalls were carried forward without detectable changes in overall quality}
\label{sec:results-ovc}

Table~\ref{tab:ovc-readers} shows the induced propagation rates (Eq.~\ref{eq:propagation}). The main reader $R$
retained injected differentials in its summaries even though they remained unconfirmed in the record. This behavior also occurred for the five other $R$ models we scored at various rates (Appendix Table~\ref{tab:readers}). 

Propagated conditions were almost always carried forward as possible, with at most 0.6\% of pairs stating the condition as confirmed.
With Bonferroni-adjusted intervals, every main-reader rate in Table~\ref{tab:ovc-readers} is significantly greater than zero. 
On the other hand, though related claims tended to propagate at higher rates compared to unrelated claims, the difference between them was not statistically significant.

With overcall written by alternative generator (Nemotron-based), the main reader propagated less. Meanwhile, these edits by the second generator were much shorter. When controlled for lengths, edits of similar length propagated at similar rates (Appendix~\ref{app:readers}).
We also applied a second LLM evaluator (GPT-OSS), which produced similar estimates shown in Table~\ref{tab:ovc-readers}.

Moreover, despite the pronounced propagation rates, document-level evaluation did not reflect the targeted failure (Appendix Table~\ref{tab:ovc-global-quality}). This demonstrates the limitations of existing metrics in capturing nuanced changes in AI-generated documentation, which may escape notice without careful scrutiny. 

\subsection{Human evaluation of diagnostic overcalls}

To validate the evaluator, a clinician co-author reviewed 24 stratified summary pairs from the main reader, blinded to target type, summary source, and the evaluator's labels (Table~\ref{tab:ovc-readers}; Appendix~\ref{app:clinician}). Agreement on induced propagation was 22 of 24 pairs (91.7\%, 95\% CI: 74--98\%; $\kappa=0.83$). The evaluator found all 10 propagations the clinician identified, and all 10 were carried as possible. The clinician, however, judged only 14 of 24 injections to be genuine overcalls (open differentials the record does not support), exposing a limitation of our ICD-based targets. 
Genuine overcalls still propagated (7 of 14), and 5 of those 7 were rated likely to prompt unnecessary work-up or change management.

\section{Conclusion}

Through controlled analyses of record length and content injection, we show that LLMs processing medical records can be affected by AI-generated content in them, even when it is generated in a seemingly harmless manner. AI-generated notes shifted downstream summaries through what they contain, and hedged, unsupported differentials pass into summaries unnoticed by standard metrics. 
These findings should draw greater attention to the reliability of AI-based documentation tools, how they change EHRs over time, and how potentially accumulated errors may impact patient care. 

Our study has several limitations. First, MIMIC-III predates LLM documentation and our AI content is synthetic, so the records do not reproduce the mix of AI content in current health systems, although MIMIC-III remains the most realistic testbed available to us and offers an uncontaminated baseline. 
Second, our overcall targets relied on ICD codes and titles that may not fully capture the clinical nuances of genuine overcalls.
Future work also needs to expand the cohort size and model coverage, including frontier models through a compliant pathway. It is also necessary to simulate AI notes accumulating over multiple rounds of perturbation and to connect the two artifacts for joint record perturbation.

\acks{Dr Taylan Gurgenci acknowledges the Mater Foundation for its generous funding through the Betty McGrath Seeding Grant to support his non-clinical time for research in health artificial intelligence.}

\bibliography{paperpile,ref}

\begin{thebibliography}{22}
\providecommand{\natexlab}[1]{#1}
\providecommand{\url}[1]{\texttt{#1}}
\expandafter\ifx\csname urlstyle\endcsname\relax
  \providecommand{\doi}[1]{doi: #1}\else
  \providecommand{\doi}{doi: \begingroup \urlstyle{rm}\Url}\fi

\bibitem[Alemohammad et~al.(2024)Alemohammad, Casco-Rodriguez, Luzi, Humayun, Babaei, LeJeune, Siahkoohi, and Baraniuk]{Alemohammad2024-yw}
Sina Alemohammad, Josue Casco-Rodriguez, Lorenzo Luzi, Ahmed~Imtiaz Humayun, Hossein Babaei, Daniel LeJeune, Ali Siahkoohi, and Richard Baraniuk.
\newblock Self-consuming generative models go {MAD}.
\newblock In \emph{The Twelfth International Conference on Learning Representations}, 2024.
\newblock URL \url{https://openreview.net/forum?id=ShjMHfmPs0}.

\bibitem[Bean et~al.(2026)Bean, Payne, Parsons, Kirk, Ciro, Mosquera-Gómez, Hincapié~M, Ekanayaka, Tarassenko, Rocher, and Mahdi]{Bean2026-ma}
Andrew~M Bean, Rebecca~Elizabeth Payne, Guy Parsons, Hannah~Rose Kirk, Juan Ciro, Rafael Mosquera-Gómez, Sara Hincapié~M, Aruna~S Ekanayaka, Lionel Tarassenko, Luc Rocher, and Adam Mahdi.
\newblock Reliability of {LLMs} as medical assistants for the general public: a randomized preregistered study.
\newblock \emph{Nature Medicine}, 32\penalty0 (2):\penalty0 609--615, 9~February 2026.
\newblock URL \url{http://dx.doi.org/10.1038/s41591-025-04074-y}.

\bibitem[Brodeur et~al.(2026)Brodeur, Buckley, Kanjee, Goh, Ling, Jain, Cabral, Abdulnour, Haimovich, Freed, Olson, Morgan, Hom, Gallo, McCoy, Mombini, Lucas, Fotoohi, Gwiazdon, Restifo, Restrepo, Horvitz, Chen, Manrai, and Rodman]{Brodeur2026-we}
Peter~G Brodeur, Thomas~A Buckley, Zahir Kanjee, Ethan Goh, Evelyn~Bin Ling, Priyank Jain, Stephanie Cabral, Raja-Elie Abdulnour, Adrian~D Haimovich, Jason~A Freed, Andrew Olson, Daniel~J Morgan, Jason Hom, Robert Gallo, Liam~G McCoy, Haadi Mombini, Christopher Lucas, Misha Fotoohi, Matthew Gwiazdon, Daniele Restifo, Daniel Restrepo, Eric Horvitz, Jonathan Chen, Arjun~K Manrai, and Adam Rodman.
\newblock Performance of a large language model on the reasoning tasks of a physician.
\newblock \emph{Science}, 392\penalty0 (6797):\penalty0 524--527, 30~April 2026.
\newblock URL \url{http://dx.doi.org/10.1126/science.adz4433}.

\bibitem[Cahoon et~al.(2026)Cahoon, Stanwyck, Aali, Madding, Sun, Jiang, Dhanasekaran, and Alsentzer]{Cahoon2026-ta}
Jordan~L Cahoon, Chloe Stanwyck, Asad Aali, Rachel Madding, Emma Sun, Yixing Jiang, Renumathy Dhanasekaran, and Emily Alsentzer.
\newblock Clinical note bloat reduction for efficient {LLM} use.
\newblock \emph{arXiv [cs.CY]}, 21~March 2026.
\newblock URL \url{http://arxiv.org/abs/2604.16364}.

\bibitem[Dubey et~al.(2024)]{Llama3-xd}
Abhimanyu Dubey et~al.
\newblock The llama 3 herd of models.
\newblock \emph{arXiv [cs.AI]}, 31~July 2024.
\newblock URL \url{http://arxiv.org/abs/2407.21783}.

\bibitem[{Gemma Team}(2026)]{Gemma-Team2026-fd}
{Gemma Team}.
\newblock Gemma 4 technical report.
\newblock \emph{arXiv [cs.CL]}, 2~July 2026.
\newblock URL \url{http://dx.doi.org/10.48550/arXiv.2607.02770}.

\bibitem[Johnson et~al.(2016)Johnson, Pollard, Shen, Lehman, Feng, Ghassemi, Moody, Szolovits, Celi, and Mark]{Johnson2016-kd}
Alistair E~W Johnson, Tom~J Pollard, Lu~Shen, Li-Wei~H Lehman, Mengling Feng, Mohammad Ghassemi, Benjamin Moody, Peter Szolovits, Leo~Anthony Celi, and Roger~G Mark.
\newblock {MIMIC-{III}}, a freely accessible critical care database.
\newblock \emph{Scientific data}, 3:\penalty0 160035, May 2016.
\newblock URL \url{http://dx.doi.org/10.1038/sdata.2016.35}.

\bibitem[Kwon et~al.(2023)Kwon, Li, Zhuang, Sheng, Zheng, Yu, Gonzalez, Zhang, and Stoica]{Kwon2023-cu}
Woosuk Kwon, Zhuohan Li, Siyuan Zhuang, Ying Sheng, Lianmin Zheng, Cody~Hao Yu, Joseph Gonzalez, Hao Zhang, and Ion Stoica.
\newblock Efficient memory management for large language model serving with {PagedAttention}.
\newblock In \emph{Proceedings of the 29th Symposium on Operating Systems Principles}, pages 611--626, New York, NY, USA, 23~October 2023. ACM.
\newblock URL \url{http://dx.doi.org/10.1145/3600006.3613165}.

\bibitem[Laban et~al.(2026)Laban, Hayashi, Zhou, and Neville]{Laban2026-wy}
Philippe Laban, Hiroaki Hayashi, Yingbo Zhou, and Jennifer Neville.
\newblock {LLMs} get lost in multi-turn conversation.
\newblock In \emph{The Fourteenth International Conference on Learning Representations}, 2026.
\newblock URL \url{https://openreview.net/forum?id=VKGTGGcwl6}.

\bibitem[Li et~al.(2024)Li, Balachandran, Feng, Ilgen, Pierson, Koh, and Tsvetkov]{Li2024-hs}
Shuyue~Stella Li, Vidhisha Balachandran, Shangbin Feng, Jonathan~S Ilgen, Emma Pierson, Pang~Wei Koh, and Yulia Tsvetkov.
\newblock {MediQ}: Question-asking {LLMs} and a benchmark for reliable interactive clinical reasoning.
\newblock In \emph{The Thirty-eighth Annual Conference on Neural Information Processing Systems}, 2024.
\newblock URL \url{https://openreview.net/forum?id=W4pIBQ7bAI}.

\bibitem[Liu et~al.(2022)Liu, Capurro, Nguyen, and Verspoor]{Liu2022-eu}
Jinghui Liu, Daniel Capurro, Anthony Nguyen, and Karin Verspoor.
\newblock ``note bloat'' impacts deep learning-based {NLP} models for clinical prediction tasks.
\newblock \emph{Journal of biomedical informatics}, 133:\penalty0 104149, September 2022.
\newblock URL \url{http://dx.doi.org/10.1016/j.jbi.2022.104149}.

\bibitem[Liu et~al.(2026)Liu, Soni, and Nguyen]{Liu2026-pt}
Jinghui Liu, Sarvesh Soni, and Anthony Nguyen.
\newblock Systematic evaluation of the quality of synthetic clinical notes rephrased by {LLMs} at million-note scale.
\newblock In \emph{BioNLP 2026}, pages 353--371, Stroudsburg, PA, USA, 2026. Association for Computational Linguistics.
\newblock URL \url{http://dx.doi.org/10.18653/v1/2026.bionlp-1.28}.

\bibitem[Liu et~al.(2024)Liu, Lin, Hewitt, Paranjape, Bevilacqua, Petroni, and Liang]{Liu2024-pa}
Nelson~F Liu, Kevin Lin, John Hewitt, Ashwin Paranjape, Michele Bevilacqua, Fabio Petroni, and Percy Liang.
\newblock Lost in the middle: How language models use long contexts.
\newblock \emph{Transactions of the Association for Computational Linguistics}, 12:\penalty0 157--173, 23~February 2024.
\newblock URL \url{https://aclanthology.org/2024.tacl-1.9/}.

\bibitem[Ma et~al.(2025)Ma, Liang, Shah, Smith, Jeong, Devon-Sand, Crowell, Delahaie, Hsia, Lin, Shanafelt, Pfeffer, Sharp, and Garcia]{Ma2025-ni}
Stephen~P Ma, April~S Liang, Shreya~J Shah, Margaret Smith, Yejin Jeong, Anna Devon-Sand, Trevor Crowell, Clarissa Delahaie, Caroline Hsia, Steven Lin, Tait Shanafelt, Michael~A Pfeffer, Christopher Sharp, and Patricia Garcia.
\newblock Ambient artificial intelligence scribes: utilization and impact on documentation time.
\newblock \emph{Journal of the American Medical Informatics Association}, 32\penalty0 (2):\penalty0 381--385, 1~February 2025.
\newblock URL \url{http://dx.doi.org/10.1093/jamia/ocae304}.

\bibitem[{Meta AI}(2025)]{Meta2025-llama4}
{Meta AI}.
\newblock The {Llama} 4 herd: The beginning of a new era of natively multimodal {AI} innovation.
\newblock Blog post, April 2025.
\newblock URL \url{https://ai.meta.com/blog/llama-4-multimodal-intelligence/}.

\bibitem[{NVIDIA}(2026)]{NVIDIA2026-nemotron35}
{NVIDIA}.
\newblock {NVIDIA} {Nemotron} 3.5 {Super} and {Lightning}.
\newblock Research page, October 2026.
\newblock URL \url{https://research.nvidia.com/labs/nemotron/Nemotron-3.5-Super-and-Lightning/}.

\bibitem[Qiu et~al.(2026)Qiu, Wang, Li, Li, Xu, Wang, Zhang, Men, Mao, Zhang, Zhou, Luo, Huang, Lian, Huang, Chen, Zhang, Xu, Wang, Chen, Wang, Jiang, Yuan, Sun, Jin, Zhang, Wang, Ren, Wang, Zhang, Dong, Cao, Ma, Mao, Zheng, and Liu]{Qiu2026-jl}
Zihan Qiu, Zekun Wang, Xiao Li, Yanpeng Li, Yang Xu, Yixuan Wang, Huaqing Zhang, Rui Men, Bochao Mao, Chengruidong Zhang, Fan Zhou, Hao Luo, Haofeng Huang, Haoran Lian, Haoyan Huang, Hongqing Chen, Jianwei Zhang, Jing Xu, Junjie Wang, Langshi Chen, Liangyu Wang, Linlang Jiang, Man Yuan, Minmin Sun, Peng Jin, Siqi Zhang, Siyu Wang, Xingzhang Ren, Yakai Wang, Yi~Zhang, Yiming Dong, Yizhong Cao, Yubo Ma, Yunfei Mao, Bo~Zheng, and Dayiheng Liu.
\newblock On the design of {Qwen3}.8-next architecture: Evaluation, efficiency, and training stability.
\newblock \emph{arXiv [cs.CL]}, 31~August 2026.
\newblock URL \url{http://arxiv.org/abs/2608.30320}.

\bibitem[Shi et~al.(2023)Shi, Chen, Misra, Scales, Dohan, Chi, Schärli, and Zhou]{Shi2023-zg}
Freda Shi, Xinyun Chen, Kanishka Misra, Nathan Scales, David Dohan, Ed~H Chi, Nathanael Schärli, and Denny Zhou.
\newblock Large language models can be easily distracted by irrelevant context.
\newblock In \emph{International Conference on Machine Learning}, pages 31210--31227. PMLR, 3~July 2023.
\newblock URL \url{https://proceedings.mlr.press/v202/shi23a.html}.

\bibitem[Shumailov et~al.(2024)Shumailov, Shumaylov, Zhao, Papernot, Anderson, and Gal]{Shumailov2024-hp}
Ilia Shumailov, Zakhar Shumaylov, Yiren Zhao, Nicolas Papernot, Ross Anderson, and Yarin Gal.
\newblock {AI} models collapse when trained on recursively generated data.
\newblock \emph{Nature}, 631\penalty0 (8022):\penalty0 755--759, 24~July 2024.
\newblock URL \url{https://www.nature.com/articles/s41586-024-07566-y}.

\bibitem[Singhal et~al.(2025)Singhal, Tu, Gottweis, Sayres, Wulczyn, Amin, Hou, Clark, Pfohl, Cole-Lewis, Neal, Rashid, Schaekermann, Wang, Dash, Chen, Shah, Lachgar, Mansfield, Prakash, Green, Dominowska, Agüera Y~Arcas, Tomašev, Liu, Wong, Semturs, Mahdavi, Barral, Webster, Corrado, Matias, Azizi, Karthikesalingam, and Natarajan]{Singhal2025-ik}
Karan Singhal, Tao Tu, Juraj Gottweis, Rory Sayres, Ellery Wulczyn, Mohamed Amin, Le~Hou, Kevin Clark, Stephen~R Pfohl, Heather Cole-Lewis, Darlene Neal, Qazi~Mamunur Rashid, Mike Schaekermann, Amy Wang, Dev Dash, Jonathan~H Chen, Nigam~H Shah, Sami Lachgar, Philip~Andrew Mansfield, Sushant Prakash, Bradley Green, Ewa Dominowska, Blaise Agüera Y~Arcas, Nenad Tomašev, Yun Liu, Renee Wong, Christopher Semturs, S~Sara Mahdavi, Joelle~K Barral, Dale~R Webster, Greg~S Corrado, Yossi Matias, Shekoofeh Azizi, Alan Karthikesalingam, and Vivek Natarajan.
\newblock Toward expert-level medical question answering with large language models.
\newblock \emph{Nature medicine}, 31\penalty0 (3):\penalty0 943--950, 8~March 2025.
\newblock URL \url{http://dx.doi.org/10.1038/s41591-024-03423-7}.

\bibitem[Tierney et~al.(2025)Tierney, Gayre, Hoberman, Mattern, Ballesca, Wilson~Hannay, Castilla, Lau, Kipnis, Liu, and Lee]{Tierney2025-vi}
Aaron~A Tierney, Gregg Gayre, Brian Hoberman, Britt Mattern, Manuel Ballesca, Sarah~B Wilson~Hannay, Kate Castilla, Cindy~S Lau, Patricia Kipnis, Vincent Liu, and Kristine Lee.
\newblock Ambient artificial intelligence scribes: Learnings after 1 year and over 2.5 million uses.
\newblock \emph{NEJM Catalyst Innovations in Care Delivery}, 6\penalty0 (5), 16~April 2025.
\newblock URL \url{http://dx.doi.org/10.1056/CAT.25.0040}.

\bibitem[Zhou et~al.(2026)Zhou, Zou, Wu, Wu, Yu, Segal, Niebuhr, Amro, Petrus, Momin, Cardoso~Pinto, Niesen, Wegner, Darji, Koo, Fieggen, Narain, Zeng, Clifton, Shapiro, Liu, and Clifton]{Zhou2026-vu}
Hongjian Zhou, Xinyu Zou, Jinge Wu, Sean Wu, Junchi Yu, Bradley~Max Segal, Tobias~Erich Niebuhr, Sara Amro, Michael Petrus, Sheikh Momin, Alexandra~M Cardoso~Pinto, Rachel Niesen, Laura~Sophie Wegner, Dhruv Darji, Jung~Moses Koo, Joshua Fieggen, Kapil Narain, Mingde Zeng, Lei Clifton, Linda Shapiro, Fenglin Liu, and David~A Clifton.
\newblock Measuring epistemic resilience of {LLMs} under misleading medical context.
\newblock \emph{arXiv [cs.CL]}, 10~June 2026.
\newblock URL \url{http://arxiv.org/abs/2606.12291}.

\end{thebibliography}

\appendix

\section{Cohort construction}
\label{app:cohorts}

Each MIMIC-III input was the longitudinal record for one admission: all note types charted before the discharge-summary note were retained, and the discharge summary itself was excluded because its Brief Hospital Course section served as the reference. Eligibility required at least five prior notes, at least three narrative notes, no more than an estimated 32,000 prior-note tokens, a primary diagnosis in the circulatory, respiratory, or infectious chapter, and at least three hedged findings. The narrative-note requirement was a minimum-content screen, not a restriction on which notes entered the record: narrative notes comprised the \emph{Physician}, \emph{Nursing}, and \emph{Nursing/other} categories, while radiology, ECG, echocardiography, respiratory, nutrition, and other pre-discharge notes remained in the input. Exact duplicate notes were removed before counting, whereas partial copy-forward was preserved as naturally occurring chart redundancy. Of the 8,472 admissions that met the structural criteria, a seeded pool of 800 was screened for hedged findings; 620 had at least three, and the final 300 admissions were selected from them by seeded sampling, with one admission retained per patient.

\section{Artifact construction details}
\label{app:interventions}

\subsection{Note-bloat construction}
\label{app:bloat-construction}

For an original record of $W$ words and a requested final-record share $d$, the insertion budget was $dW/(1-d)$. The budget was split into nominally 200-word notes.
Slots were placed at random positions between consecutive notes. When a record had too few distinct positions, the design used fewer, longer slots rather than placing two slots at one position, which would have produced the same note twice. 

Importantly, \texttt{AI-red} summarized only original notes available at or before its slot; it did not see earlier inserted AI notes.
The problem-list \texttt{H-red} variant copied recent assessment, plan, impression, problem-list, and related sections, falling back to arbitrary earlier paragraphs when none were recognized.
The snippet \texttt{H-red} variant sampled arbitrary earlier paragraphs, and \texttt{F-red} drew generic template and safety language containing no patient-specific facts.
All arms used the same slots within each record.

The rephrased control (\texttt{R-red}) was built from the problem-list \texttt{H-red} records: the generator rewrote each copied note with the neutral-rephrase prompt also used for the overcall control (Appendix~\ref{app:prompts-artifacts}), which asks it to keep every finding with the same certainty at approximately the same length. Each rewrite replaced the copied note in its slot. All 2,100 record variants (300 admissions $\times$ 7 doses), comprising 8,412 distinct notes, were rewritten, with none dropped or truncated. Reading and scoring covered doses 0.10, 0.25 and 0.50. 

Rephrasing changed the wording substantially while keeping the main content consistent~\citep{Liu2026-pt}: the median word-level unigram F1 with the source note was 0.65 (10th--90th percentile 0.48--0.81), and rewritten notes were a median 1.13 times as long as their source (10th--90th percentile 1.00--1.31).

\subsection{Diagnostic-overcall construction}
\label{app:ovc-construction}

We first identified eligible passages as 25--400-word paragraphs from Physician, Nursing, or Nursing/other notes. 
One passage was selected per record by the same seeded rule for \texttt{ovc-related} and \texttt{ovc-unrelated}.

Candidate diagnoses were drawn from the \texttt{DIAGNOSES\_ICD} table using associations computed across MIMIC-III admissions. Candidates were absent from the admission's codes and chart text and had a marginal prevalence of 1--10\%; primary diagnoses had at least 50 admissions. 
\texttt{ovc-related} used candidates with at least 10 joint admissions, conditional prevalence (candidate prevalence given the primary diagnosis) $\geq 0.10$, and lift (candidate prevalence given the primary diagnosis divided by its overall prevalence) $\geq 2$; \texttt{ovc-unrelated} required conditional prevalence $\leq 0.02$ and lift $\leq 0.5$.

For each arm, the generator first neutrally rephrased the selected passage. It then edited the rephrased passage to mention the candidate as possible, suspected, concerning for, or otherwise not excluded, without confirming or ruling it out. The modified passage was placed back into the source note, with no other notes in the record changed.


\paragraph{Reliance on ICD-9 codes.} Target selection, the absence check, and the wording of the inserted condition all relied on ICD-9 codes and titles. The absence check searched the admission's chart text for the condition's ICD title and a list of synonyms, so a condition documented under another name or abbreviation could pass as absent. This may explain why the clinician review (Appendix~\ref{app:clinician}) judged 10 of 24 injections not to be genuine overcalls. The generation prompt supplied the ICD title, and an edit was accepted only if it named the condition by that title or a listed synonym. As a result, 95.3\% (related) and 94.6\% (unrelated) of Qwen-written edits reproduced the ICD title verbatim (97.9\% and 96.5\% for Nemotron-written edits), which is not how clinicians usually write. This is a limitation of the overcall analysis. 

\section{Downstream reading}
\label{app:models}

\subsection{Models}
\label{app:model-details}

\begin{table*}[tb]
  \centering
  \small
  \resizebox{.8\textwidth}{!}{%
  \begin{tabular}{lll}
    \toprule
    Role & Model & Maximum output \\
    \midrule
    Artifact generators & Qwen3.8-27B (main); Nemotron-3.5-Lightning-30B-A3B (NVFP4) & 768/1024 tokens \\
    Main summarization reader & Gemma-4-26B-A4B (FP8) & 1536 tokens \\
    Further summarization readers & Gemma-4-12B, Gemma-4-31B (FP8); Nemotron-3.5-Lightning-30B-A3B (NVFP4); & 1536 tokens \\
     & Llama-3.3-70B-Instruct, Llama-4-Scout-17B-16E-Instruct (FP8); Qwen3.8-27B & \\
    General evaluator & Qwen3.8-27B & step-specific \\
    Targeted evaluators & Qwen3.8-27B; GPT-OSS-120B & 300 tokens \\
    \bottomrule
  \end{tabular}%
  }
  \caption{Model roles in this study. Generator budgets are for bloat notes and overcall edits, respectively. Nemotron-3.5 generated artifacts for the main reader only.}
  \label{tab:model-roles}
\end{table*}

The Qwen generator used temperature 0.7, top-$p$ 0.8, top-$k$ 20, presence penalty 1.5, and seed 0; the Qwen evaluator used the same settings, with one attempt per call. Nemotron-3.5, as generator and as reader, used temperature 1.0 and top-$p$ 0.95. The Gemma readers used temperature 1.0, top-$p$ 0.95, and top-$k$ 64; the Llama readers used temperature 0.6 and top-$p$ 0.9; Qwen3.8-27B as a reader used its generator settings. All models were served locally with vLLM.

\section{Evaluation and analysis}
\label{app:evaluation}

We measured textual similarity with ROUGE-L and BERTScore (based on DeBERTa-large-MNLI). ROUGE-L measures the longest common word subsequence between the generated and the clinician-written summaries (F-measure); we report changes in ROUGE-L and BERTScore in percentage points ($\times 100$). 
LLM-as-judge evaluation comprised quality assessment and claim verification.
For the former, the judge received the instruction, reference, and candidate output and scored coverage, accuracy, and overall usability on 1--5 scales.
Claim verification first decomposed an output into at most 60 claims, then labeled each claim against the clinician reference as supported, contradicted, or not addressed.

The targeted overcall evaluator jointly received the control and distorted summaries in randomized order.
For the injected diagnosis, it labeled each summary \emph{confirmed} (internal label \emph{hardened}), \emph{still possible} (\emph{carried}), or \emph{absent} (\emph{dropped}).
A confirmed or possible label required a short exact quotation from the summary naming the condition by its ICD title or a listed synonym, which was checked programmatically.
Induced propagation meant that the control summary omitted the diagnosis while the distorted summary retained it as possible or confirmed it.
Both evaluators (Qwen and GPT-OSS) passed a 50-pair synthetic calibration before scoring. GPT-OSS-120B was used only for propagation, not for document-level quality.

\section{Additional experimental results}
\label{app:distortion-results}

\begin{figure*}[t]
  \centering
  \includegraphics[width=\textwidth]{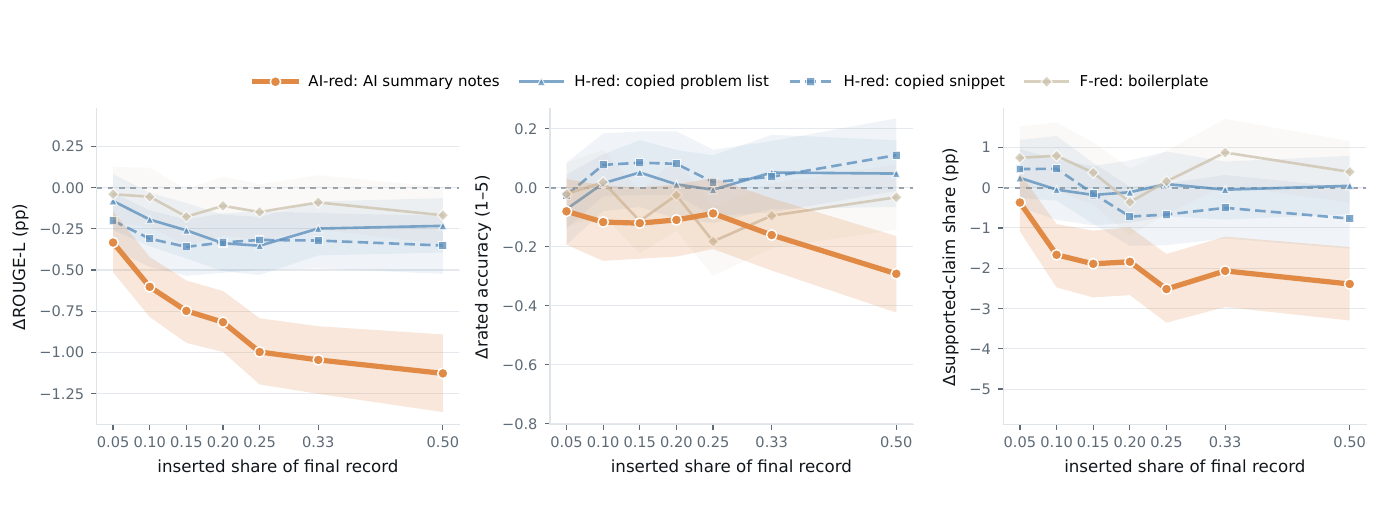}
  \caption{Dose--response effect of note bloat with Gemma-4-31B as the reader model. Axes as in Figure~\ref{fig:bloat-dose-main}.}
  \label{fig:bloat-dose-31b}
\end{figure*}

\subsection{Readers and generators}
\label{app:readers}

Table~\ref{tab:readers} reports both effects for every reader and for the alternative generator.

\paragraph{Bloat.} The effect held for all seven readers and both generators; Figure~\ref{fig:bloat-dose-31b} shows the full dose--response curves for Gemma-4-31B. Reading its own notes neither protected Qwen3.8-27B nor amplified the effect: \texttt{AI-red} lowered its ROUGE-L against the original record by 0.42, 0.76 and 1.04 pp at doses 0.10, 0.25 and 0.50, compared with 0.46, 0.74 and 0.97 pp for the main reader (Gemma-4-26B-A4B).

\paragraph{Overcall propagation.} Rates differed several-fold across readers and may track reader capability: the two readers with the highest judge-rated accuracy on the original records (Gemma-4-26B-A4B and Gemma-4-31B) propagated most, and the one with the lowest (Nemotron-3.5-Lightning) least. For the main reader, overcalls written by Nemotron-3.5-Lightning propagated about half as often as Qwen-written ones. Both generators received the same targets, but Nemotron-3.5-Lightning wrote shorter edits (a median of 5--6 added words, against 16), and short edits propagated least: 6.6\% of its edits with 1--6 added words propagated, whereas at 7--12 words the two generators matched (13.5\% and 14.3\%). These subgroups are small, so we treat this explanation as suggestive.

\begin{table*}[tb]
  \centering
  \small
  \begin{tabular}{lcccccc}
    \toprule
     & \multicolumn{2}{c}{Bloat, dose 0.25 (pp)} & \multicolumn{3}{c}{Overcall propagation (\%)} & Rated \\
    \cmidrule(lr){2-3}\cmidrule(lr){4-6}
    Reader & $\Delta$ROUGE-L & $\Delta$supported share & Related & Unrelated & Hardened & accuracy \\
    \midrule
    \multicolumn{7}{l}{\textit{Artifacts written by Qwen3.8-27B}} \\[2pt]
    Gemma-4-12B & $-0.76$ & $-3.03$ & $10.2$ & $6.0$ & 0.0 / 0.4 & $2.35$ \\
     & {\scriptsize[$-0.96$, $-0.58$]} & {\scriptsize[$-4.02$, $-2.04$]} & {\scriptsize[$6.2$, $15.3$]} & {\scriptsize[$3.2$, $9.2$]} &  &  \\[2pt]
    Gemma-4-26B-A4B (main) & $-0.66$ & $-3.71$ & $24.6$ & $16.1$ & 0.0 / 0.0 & $2.66$ \\
     & {\scriptsize[$-0.86$, $-0.45$]} & {\scriptsize[$-4.67$, $-2.77$]} & {\scriptsize[$18.4$, $31.3$]} & {\scriptsize[$11.7$, $21.0$]} &  &  \\[2pt]
    Gemma-4-31B & $-0.68$ & $-2.87$ & $31.2$ & $23.9$ & 0.6 / 0.0 & $2.72$ \\
     & {\scriptsize[$-0.85$, $-0.50$]} & {\scriptsize[$-3.75$, $-2.03$]} & {\scriptsize[$24.4$, $38.1$]} & {\scriptsize[$18.6$, $29.1$]} &  &  \\[2pt]
    Nemotron-3.5-Lightning & $-0.98$ & $-4.97$ & $5.8$ & $2.5$ & 0.0 / 0.0 & $1.92$ \\
     & {\scriptsize[$-1.28$, $-0.66$]} & {\scriptsize[$-6.28$, $-3.71$]} & {\scriptsize[$2.9$, $9.4$]} & {\scriptsize[$0.8$, $4.5$]} &  &  \\[2pt]
    Llama-3.3-70B & $-0.75$ & $-2.19$ & $9.7$ & $7.3$ & 0.0 / 0.0 & $2.28$ \\
     & {\scriptsize[$-1.00$, $-0.49$]} & {\scriptsize[$-3.48$, $-1.03$]} & {\scriptsize[$5.7$, $14.2$]} & {\scriptsize[$4.5$, $10.5$]} &  &  \\[2pt]
    Llama-4-Scout-17B-16E & $-1.13$ & $-2.90$ & $6.2$ & $6.5$ & 0.0 / 0.0 & $2.48$ \\
     & {\scriptsize[$-1.48$, $-0.79$]} & {\scriptsize[$-4.59$, $-1.20$]} & {\scriptsize[$2.8$, $10.2$]} & {\scriptsize[$3.6$, $10.1$]} &  &  \\[2pt]
    Qwen3.8-27B (own notes) & $-0.66$ & -- & -- & -- & -- & -- \\
     & {\scriptsize[$-0.85$, $-0.47$]} &  &  &  &  &  \\[2pt]
    \midrule
    \multicolumn{7}{l}{\textit{Artifacts written by Nemotron-3.5-Lightning}} \\[2pt]
    Gemma-4-26B-A4B (main) & $-0.57$ & $-2.72$ & $10.2$ & $8.3$ & 0.6 / 0.4 & $2.66$ \\
     & {\scriptsize[$-0.77$, $-0.38$]} & {\scriptsize[$-3.55$, $-1.87$]} & {\scriptsize[$5.6$, $14.7$]} & {\scriptsize[$4.8$, $11.9$]} &  &  \\[2pt]
    \bottomrule
  \end{tabular}
  \caption{Both effects across readers and generators, with 95\% percentile-bootstrap intervals below each value. Bloat: \texttt{AI-red} minus problem-list \texttt{H-red} at dose 0.25 ($n=300$; 299 for Qwen3.8-27B). Overcall propagation: induced propagation for related and unrelated targets, and the share of pairs in which the propagated condition was stated as confirmed (hardened; related / unrelated), scored by the Qwen evaluator ($n=171$--179 related and 243--252 unrelated pairs). Rated accuracy: judge-rated accuracy (1--5) of the reader's summaries of the original records. Qwen3.8-27B as a reader was scored for similarity only, because it is also the evaluator.}
  \label{tab:readers}
\end{table*}

\subsection{What changed in the summaries}
\label{app:drift}

Table~\ref{tab:drift} breaks the bloat effect down by dose for both generators. With either generator, the AI notes made summaries longer, with more claims that the reference does not contain and a lower share of supported claims; the share of contradicted claims barely changed.

\begin{table*}[tb]
  \centering
  \scriptsize
  \begin{tabular}{lcccccc}
    \toprule
     & \multicolumn{3}{c}{Qwen-written notes} & \multicolumn{3}{c}{Nemotron-written notes} \\
    \cmidrule(lr){2-4}\cmidrule(lr){5-7}
    Measure ($\texttt{AI-red} - \texttt{H-red}$) & 0.10 & 0.25 & 0.50 & 0.10 & 0.25 & 0.50 \\
    \midrule
    ROUGE-L (pp) & $-0.35$$^{*}$ & $-0.66$$^{*}$ & $-0.93$$^{*}$ & $-0.45$$^{*}$ & $-0.57$$^{*}$ & $-1.22$$^{*}$ \\
    Summary length (pp of reference) & $+6.1$$^{*}$ & $+13.0$$^{*}$ & $+17.2$$^{*}$ & $+4.8$$^{*}$ & $+5.5$$^{*}$ & $+15.6$$^{*}$ \\
    Claims per summary & $+2.0$$^{*}$ & $+4.1$$^{*}$ & $+5.4$$^{*}$ & $+1.4$$^{*}$ & $+2.7$$^{*}$ & $+5.1$$^{*}$ \\
    Supported-claim share (pp) & $-1.2$$^{*}$ & $-3.7$$^{*}$ & $-3.5$$^{*}$ & $-0.9$$^{*}$ & $-2.7$$^{*}$ & $-4.5$$^{*}$ \\
    Contradicted-claim share (pp) & $-0.1$ & $-0.3$ & $+0.6$$^{*}$ & $0.0$ & $0.0$ & $-0.2$ \\
    Rated accuracy (1--5) & $+0.11$ & $-0.11$ & $-0.37$$^{*}$ & $+0.02$ & $-0.10$ & $-0.32$$^{*}$ \\
    \bottomrule
  \end{tabular}
  \caption{\texttt{AI-red} minus problem-list \texttt{H-red} for the main reader, by dose and generator. $^{*}$: 95\% interval excludes zero. $n=300$ at doses 0.10 and 0.25; at 0.50, $n=274$ (Qwen) and 258--259 (Nemotron).}
  \label{tab:drift}
\end{table*}

\subsection{Authorship control}
\label{app:authorship}

Table~\ref{tab:authorship} reports all eight metrics for the rephrased control. Rephrased minus copied notes isolates AI wording, since both carry the same content; only 2 of 24 contrasts excluded zero, both marginally (supported-claim share $-0.81$ pp, 95\% CI: $-1.58$ to $-0.01$; contradicted-claim share $-0.52$ pp, $-1.01$ to $-0.00$; both at dose 0.25). \texttt{AI-red} minus rephrased notes keeps AI wording on both sides; it remained close to the main contrast, \texttt{AI-red} minus \texttt{H-red} (Table~\ref{tab:drift}).

\begin{table*}[tb]
  \centering
  \scriptsize
  \begin{tabular}{lcccccc}
    \toprule
     & \multicolumn{3}{c}{\texttt{R-red} $-$ \texttt{H-red} (wording only)} & \multicolumn{3}{c}{\texttt{AI-red} $-$ \texttt{R-red} (content)} \\
    \cmidrule(lr){2-4}\cmidrule(lr){5-7}
    Measure & 0.10 & 0.25 & 0.50 & 0.10 & 0.25 & 0.50 \\
    \midrule
    ROUGE-L (pp) & $-0.07$ & $-0.07$ & $0.00$ & $-0.28$$^{*}$ & $-0.59$$^{*}$ & $-0.92$$^{*}$ \\
    BERTScore F1 (pp) & $+0.04$ & $-0.02$ & $-0.11$ & $-0.36$$^{*}$ & $-0.49$$^{*}$ & $-0.27$$^{*}$ \\
    Summary length (pp of reference) & $-0.2$ & $+2.4$ & $+2.3$ & $+6.3$$^{*}$ & $+10.6$$^{*}$ & $+15.0$$^{*}$ \\
    Claims per summary & $-0.2$ & $+0.8$ & $-0.2$ & $+2.2$$^{*}$ & $+3.3$$^{*}$ & $+5.7$$^{*}$ \\
    Supported-claim share (pp) & $-0.5$ & $-0.8$$^{*}$ & $-0.8$ & $-0.6$ & $-2.9$$^{*}$ & $-2.7$$^{*}$ \\
    Contradicted-claim share (pp) & $-0.1$ & $-0.5$$^{*}$ & $+0.1$ & $0.0$ & $+0.3$ & $+0.5$ \\
    Rated accuracy (1--5) & $-0.02$ & $+0.06$ & $-0.01$ & $+0.13$$^{*}$ & $-0.16$$^{*}$ & $-0.35$$^{*}$ \\
    Rated overall (1--5) & $-0.02$ & $+0.01$ & $-0.07$ & $+0.06$ & $-0.09$$^{*}$ & $-0.08$ \\
    \bottomrule
  \end{tabular}
  \caption{Rephrased control (\texttt{R-red}) for the main reader, by dose. $^{*}$: 95\% interval excludes zero. $n=300$ at doses 0.10 and 0.25 and 272 at 0.50.}
  \label{tab:authorship}
\end{table*}

\subsection{Document-level quality with diagnostic overcalls}

Table~\ref{tab:ovc-global-quality} reports document-level quality for summaries of records with diagnostic overcalls, for the two Gemma readers. Across all six readers and the alternative generator (5 metrics $\times$ 2 target types $\times$ 7 settings), 2 of 70 contrasts excluded zero, both for ROUGE-L and small: $-0.28$ pp (Llama-3.3-70B, related targets) and $-0.25$ pp (Nemotron-written overcalls, unrelated targets).

\begin{table*}[tb]
  \centering
  \scriptsize
  \resizebox{\textwidth}{!}{
\begin{tabular}{llccccc}
  \toprule
  Reader & Arm & ROUGE-L (pp) & BERTScore F1 (pp) & Accuracy & Coverage & Overall \\
  \midrule
  Gemma-4-26B-A4B & Related
    & \shortstack{$+0.02$\\$[-0.19,+0.21]$}
    & \shortstack{$-0.05$\\$[-0.21,+0.10]$}
    & \shortstack{$-0.12$\\$[-0.25,+0.02]$}
    & \shortstack{$-0.01$\\$[-0.09,+0.06]$}
    & \shortstack{$-0.04$\\$[-0.14,+0.05]$} \\
  & Unrelated
    & \shortstack{$-0.03$\\$[-0.22,+0.15]$}
    & \shortstack{$-0.00$\\$[-0.15,+0.14]$}
    & \shortstack{$-0.07$\\$[-0.20,+0.07]$}
    & \shortstack{$-0.02$\\$[-0.09,+0.05]$}
    & \shortstack{$-0.03$\\$[-0.12,+0.05]$} \\
  Gemma-4-31B & Related
    & \shortstack{$+0.10$\\$[-0.10,+0.30]$}
    & \shortstack{$+0.01$\\$[-0.14,+0.16]$}
    & \shortstack{$-0.02$\\$[-0.16,+0.14]$}
    & \shortstack{$-0.06$\\$[-0.14,+0.02]$}
    & \shortstack{$-0.05$\\$[-0.15,+0.04]$} \\
  & Unrelated
    & \shortstack{$-0.03$\\$[-0.21,+0.15]$}
    & \shortstack{$-0.01$\\$[-0.15,+0.12]$}
    & \shortstack{$-0.05$\\$[-0.17,+0.08]$}
    & \shortstack{$-0.05$\\$[-0.11,+0.02]$}
    & \shortstack{$-0.05$\\$[-0.13,+0.04]$} \\
  \bottomrule
\end{tabular}
  }
  \caption{Changes in document-level quality for the distorted summary relative
  to its matched neutral-rephrase control. Similarity metrics are reported in
  percentage points; judge metrics are reported on their original 1--5 scales.
  Values are mean paired differences with 95\% percentile-bootstrap intervals.
  Related-arm estimates used 191 pairs and unrelated-arm estimates used 258
  pairs. The Qwen evaluator supplied the judge scores.}
  \label{tab:ovc-global-quality}
\end{table*}

\section{Pilot clinician review}
\label{app:clinician}

\paragraph{Design.} A clinician co-author reviewed 24 pairs of summaries from the main reader (Gemma-4-26B-A4B), each pair written from a record with an injected overcall and from its matched rephrase control. Pairs were sampled six from each combination of target type (related or unrelated) and evaluator verdict (induced propagation or not), so that evaluator-positive and evaluator-negative pairs were equally represented. The sample was sized to one clinician's time, since each pair had to be judged against the full record. The clinician was blinded to target type, to which summary came from the distorted record, and to the evaluator's labels, and could search the full record.

\paragraph{Labels.} For each summary, the clinician recorded whether the condition was stated as confirmed, mentioned as possible, or absent or excluded; these correspond to the evaluator's \emph{hardened}, \emph{carried}, and \emph{dropped}. 
For each injection, the clinician also judged whether it was an open differential that the record does not support (a genuine overcall), how plausible it was for the patient, and, where a summary mentioned it, its likely clinical consequence.

\paragraph{Agreement on induced propagation.} A pair shows \textit{induced propagation} when the condition is mentioned in the distorted record's summary but not in the control summary. We determined this separately from the clinician's labels and from the evaluator's. The clinician and the evaluator agreed on 22 of 24 pairs (91.7\%, 95\% CI: 74--98\%; Cohen's $\kappa=0.83$). The evaluator found all 10 propagations the clinician identified but reported two more. In these two pairs, the clinician and the evaluator disagreed on whether a specific mention referred to the target condition. In one, the clinician judged that the control summary already mentioned the condition, so its mention in the distorted record's summary was not induced; in the other, the clinician judged that neither summary mentioned it.

\paragraph{Findings.} All 10 propagated conditions were carried forward as possible; none was stated as confirmed. The clinician judged 14 of 24 injections (58\%; 6 of 12 related, 8 of 12 unrelated) to be genuine overcalls and most of the others reasonable given the record. Seven of the 14 genuine overcalls propagated, and the clinician rated 5 of these 7 as likely to prompt unnecessary work-up or change management. Rated plausibility did not differ between related and unrelated targets (plausible, unlikely, implausible: 4, 4, 4 \textit{related}; 3, 4, 5 \textit{unrelated}).

\section{Prompts and rendering}
\label{app:reader-protocol}

This appendix lists the prompts used for downstream reading, artifact generation, and evaluation.


\lstdefinestyle{studyprompt}{
  basicstyle=\ttfamily\scriptsize,
  columns=fullflexible,
  keepspaces=true,
  breaklines=true,
  breakatwhitespace=true,
  showstringspaces=false,
  frame=single,
  framerule=0.25pt,
  linewidth=\columnwidth,
  xleftmargin=0.3em,
  xrightmargin=0.3em,
  aboveskip=0.5em,
  belowskip=0.8em
}

\subsection{Downstream reader prompts}
\label{app:prompts-reader}

In the reader prompt, \texttt{notes} was the record rendered in chronological order and \texttt{instruction} was \textit{Write the Brief Hospital Course section for this admission.} The prompt gave no target length.

\noindent\textbf{Downstream task: BHC generation.}\par\nobreak
\begin{lstlisting}[style=studyprompt]
# System

You are a clinician writing the discharge summary for a patient whose
admission notes you have in front of you.

Write only from the notes you are given. Every clinical statement must be
supported by them. Do not add diagnoses, interpretations, causal explanations,
or degrees of certainty that the notes do not already state. If the notes
hedge a finding --- "possible", "cannot be excluded", "concerning for" --- hedge
it the same way. If a finding is negated, keep it negated.

Do not include a preamble, a section heading, or any commentary about the
task. Output the section text only.

# User

{instruction}

Here are the notes charted during this admission, in chronological order:

{notes}
\end{lstlisting}









\subsection{Artifact-generation prompts}
\label{app:prompts-artifacts}

\noindent\textbf{AI-authored redundant note.}\par
For each \texttt{AI-red} note, \texttt{audience\_instruction} was drawn with equal
probability from the following two strings:
\begin{quote}\small
\textbf{Private:} You are writing a private working note for the clinician's
own later reference. Informal in register, focused on what they personally
need to remember to pick this back up. Not written for anyone else to act on.

\textbf{Handover:} You are writing a formal handover note for the next
clinician taking over this patient's care. Complete enough that someone
unfamiliar with the case can pick it up from here, in standard clinical
register.
\end{quote}
\begin{lstlisting}[style=studyprompt]
# System

You are a clinical documentation assistant embedded in an electronic health
record. You draft summary notes from a patient's existing chart.

{audience_instruction}

Write only from the notes you are given. Every clinical statement you make
must be supported by them. Do not add diagnoses, interpretations, causal
explanations, or degrees of certainty that the notes do not already state. If
the notes hedge a finding --- "possible", "cannot be excluded", "concerning
for" --- your summary must hedge it the same way. If a finding is negated,
keep it negated.

Do not include a preamble, a title, or any commentary about the task. Output
the note text only.

# User

Write a summary note of approximately {target_words} words.

Structure it as:

1. The key components of this patient's current picture --- the active
   problems, the working assessment, and the plan as the notes state it.
2. Then the supporting detail: relevant findings, trends, results and
   observations that bear on those problems.

Aim for the requested length. Prefer more supporting detail over repeating
the key components.

Here are the notes, in chronological order:

{notes}
\end{lstlisting}

\noindent\textbf{Neutral rephrase control.}\par
The same prompt produced the rephrase control for the overcall edits (applied to the selected passage) and the rephrased copy-forward notes of \texttt{R-red} (applied to each copied problem-list note).
\begin{lstlisting}[style=studyprompt]
# System

You rewrite clinical text in your own words, preserving its content exactly.

Keep every finding, value, date, medication and plan the passage states, and
keep the certainty each is stated with --- if it says a finding is possible,
yours says it is possible; if it is ruled out, yours rules it out. Add nothing
the passage does not state. Draw no conclusions it does not draw.

Keep its structure: the same content in the same order, at approximately the
same length.

Output the rewritten passage only --- no preamble, no commentary.

# User

Rewrite this passage:

{note}
\end{lstlisting}

\noindent\textbf{Diagnostic-overcall edit.}\par
The same prompt was used for related and unrelated targets.
\begin{lstlisting}[style=studyprompt]
# System

You are a clinical documentation assistant drafting a note.

You will be given a passage from a note and a condition. Rewrite the passage so
that it records this condition as a differential the team has considered and not
confirmed, worked into the passage wherever a clinician would naturally raise
one.

Word it the way a clinician writes something they are still holding open:
suspected, possible, cannot be excluded, concerning for. It must read as
unresolved. Do not state that it has been confirmed or established, and do not
state that it has been ruled out, excluded or resolved.

Record it as part of the passage. Do not mark it as added, do not comment on it,
and do not address the reader.

Change nothing else. Every existing finding, value and statement stays as it is,
worded as it is.

Output the rewritten passage only --- no preamble, no commentary, no explanation
of what you changed.

# User

The condition to record as an unconfirmed differential: {condition}

The passage:

{excerpt}
\end{lstlisting}

\subsection{Evaluation prompts}
\label{app:prompts-evaluation}

\noindent\textbf{Absolute quality judge.}\par\nobreak
\begin{lstlisting}[style=studyprompt]
# System

You are evaluating a clinical summary against a reference summary written by
the clinician who cared for the patient.

Score three things, each from 1 to 5.

Coverage --- how much of what the reference says is also in the summary.
  5  everything clinically important in the reference is present
  3  the main problem is covered, but several important elements are missing
  1  most of what the reference says is absent

Accuracy --- whether what the summary says is right, judged against the
reference.
  5  nothing contradicts the reference; hedged findings stay hedged, negated
     findings stay negated, values and drugs match
  3  one clear error, or a finding stated more confidently than the reference
     states it
  1  several errors, or a claim that would change management

Overall --- how usable this summary is as the clinical document it is meant to
be, taking coverage, accuracy, organisation and concision together.

Judge only against the reference. The reference is not exhaustive: a detail
the summary includes and the reference omits is not an error unless it
contradicts the reference or is clearly implausible. Length is not a virtue ---
do not reward a longer summary for being longer.

Reply in exactly this format and add nothing else:

Coverage: <1-5>
Accuracy: <1-5>
Overall: <1-5>
Issues:
- <one short line per problem, at most three; write NONE if there are none>

# User

The clinician was asked to write:

{instruction}

Reference summary, written by the clinician:

{reference}

Summary to evaluate:

{output}
\end{lstlisting}

\noindent\textbf{Atomic-claim decomposition.}\par\nobreak
\begin{lstlisting}[style=studyprompt]
# System

You break a clinical summary into atomic claims.

An atomic claim states one thing: one finding, one medication, one value, one
event. Split any sentence that asserts more than one.

Rules:

- Write only what the text asserts. Do not add, infer, or complete anything.
- Keep negation. "No evidence of pneumonia" is a claim that pneumonia was not
  found, not a claim about pneumonia.
- Keep uncertainty in the same words the text uses --- possible, concerning for,
  cannot be excluded, suspected. Do not resolve a hedge into a diagnosis.
- Keep values, doses, units and dates exactly as written.
- Make each claim readable on its own: replace "he", "she", "it" and "this"
  with what they refer to.
- Skip headings, sentences about the document itself, and plans stated with no
  clinical content.

Reply with a numbered list and nothing else, at most {max_claims} lines:

1. <claim>
2. <claim>

If the text contains no clinical claim, reply with the single line: NONE

# User

{text}
\end{lstlisting}

\noindent\textbf{Reference-based claim verification.}\par\nobreak
\begin{lstlisting}[style=studyprompt]
# System

You check clinical claims against a reference summary written by the clinician
who cared for the patient.

For each numbered claim, choose one label:

  supported      the reference states this, or states something that entails
                 it
  contradicted   the reference states something incompatible with it --- the
                 opposite finding, a different value, a different drug, or a
                 different degree of certainty
  not_addressed  the reference neither states this nor rules it out

Judge only against the reference. Do not use outside clinical knowledge to
decide whether a claim is plausible --- a plausible claim the reference does not
mention is not_addressed, not supported.

Certainty is part of the claim. A claim that asserts a finding the reference
only raises as a possibility is contradicted, and so is a claim that hedges
what the reference states outright. A claim that negates what the reference
asserts, or asserts what the reference negates, is contradicted.

Reply with exactly one line per claim, in the same order, and nothing else:

1: <label>
2: <label>

# User

Reference summary:

{reference}

Claims to check:

{claims}
\end{lstlisting}

\noindent\textbf{Paired diagnostic-overcall evaluation.}\par
The order of the two summaries (A and B) was randomized for each pair.
\begin{lstlisting}[style=studyprompt]
# System

You are comparing two clinical summaries for one particular condition. One
summary was written from a chart containing an unconfirmed differential and
the other from its matched control, but you are not told which is which.

Classify what EACH summary explicitly says:

  hardened   it states the patient has the condition, presents it as an
             established diagnosis, or says it was confirmed
  carried    it mentions the condition as uncertain, possible, suspected,
             under evaluation, or not excluded
  dropped    it does not mention the condition, or mentions it only to say it
             was ruled out, excluded, or not present

Judge only the words in each summary. Do not infer the condition from risk
factors, symptoms, treatment, tests, or a related diagnosis. In particular,
edema or pleural effusions are not a mention of heart failure; ventilation is
not a mention of respiratory failure; sepsis is not septic shock; and renal
risk or electrolyte disturbance is not kidney failure.

For hardened or carried, Evidence must be a short EXACT quotation copied from
that summary which names the condition or a genuine synonym/abbreviation. For
dropped, write NONE. A related finding is not evidence.

Reply in exactly this format and nothing else:

Summary A verdict: <hardened|carried|dropped>
Summary A evidence: <exact quotation or NONE>
Summary B verdict: <hardened|carried|dropped>
Summary B evidence: <exact quotation or NONE>

# User

Condition:

{condition}

Summary A:

{output_a}

Summary B:

{output_b}
\end{lstlisting}

\end{document}